\documentclass[twoside,11pt]{article}

\usepackage{blindtext}
\usepackage{listings}

\usepackage{xcolor}
\usepackage{wrapfig}
\usepackage{tcolorbox}
\usepackage{rotating}  
\usepackage{placeins} 

\usepackage{pythonhighlight}
\usepackage{tabularx}
\usepackage{array}
\usepackage{booktabs}
\usepackage{xcolor}
\usepackage{colortbl}

\definecolor{codegreen}{rgb}{0,0.6,0}
\definecolor{codegray}{rgb}{0.5,0.5,0.5}
\definecolor{codepurple}{rgb}{0.58,0,0.82}
\definecolor{backcolour}{rgb}{0.95,0.95,0.92}

\lstdefinestyle{mystyle}{
    backgroundcolor=\color{backcolour},   
    commentstyle=\color{codegreen},
    keywordstyle=\color{magenta},
    numberstyle=\tiny\color{codegray},
    stringstyle=\color{codepurple},
    basicstyle=\ttfamily\footnotesize,
    breakatwhitespace=false,         
    breaklines=true,                 
    captionpos=b,                    
    keepspaces=true,                 
    numbers=left,                    
    numbersep=5pt,                  
    showspaces=false,                
    showstringspaces=false,
    showtabs=false,                  
    tabsize=2,
}

\usepackage[preprint]{jmlr2e}

\usepackage[skip=1pt, indent=16pt]{parskip}

\usepackage{lastpage}
\jmlrheading{23}{2026}{1-\pageref{LastPage}}{1/21; Revised 5/22}{9/22}{21-0000}{Joseph D. Viviano and Salem Lahlou}

\ShortHeadings{\texttt{torchgfn: a PyTorch gflownet library}}{Viviano, Younis, Choi, Schmidt, Bengio, \& Lahlou}
\firstpageno{1}
\usepackage{cleveref}
\crefname{section}{Section}{Sections}
\crefname{appendix}{Appendix}{Appendices}
\crefname{subsection}{Section}{Sections}
\crefname{subsubsection}{Section}{Sections}
\crefname{table}{Table}{Tables}

\usepackage{cleveref}

\begin{document}

\title{\texttt{torchgfn}: A PyTorch GFlowNet library}

\author{Joseph D. Viviano*$^{\alpha}$, Omar G. Younis*$^{\alpha}$, Sanghyeok Choi*$^{\alpha,\eta}$, Victor Schmidt$^{\alpha,\beta}$, Yoshua Bengio$^{\alpha,\gamma,\delta}$, \and Salem Lahlou$^{\zeta,\omega}$ \\
\email joseph@viviano.ca, omar.younis@mila.quebec, sanghyeok.choi@ed.ac.uk, \\
salem.lahlou@mbzuai.ac.ae
} 

\maketitle
\vspace{-7mm}
\noindent \small{$^\alpha$Mila, $^\beta$Entalpic, $^\gamma$Université de Montréal, $^\delta$CIFAR,, $^\zeta$MBZUAI, $^{\eta}$University of Edinburgh, $^{\omega}$Work started at Mila. $^*$denotes co-first authorship }\footnote{ordered by geographic proximity to Saint Joseph's Oratory of Mount Royal at the time of publication. Co-first authors agree that they can list their names in any order on their CVs.} 

\vspace{1mm}
\begin{abstract}
The growing popularity of generative flow networks (GFlowNets or GFNs) among a range of researchers with diverse backgrounds and areas of expertise necessitates a library that facilitates the testing of new features (\textit{e.g.}, training losses and training policies) against standard benchmark implementations, or on a set of common environments. We present \texttt{torchgfn}, a PyTorch library that aims to address this need. Its core contribution is a modular and decoupled architecture which treats environments, neural network modules, and training objectives as interchangeable components. This provides users with a simple yet powerful API to facilitate rapid prototyping and novel research. Multiple examples are provided, replicating and unifying published results. 
The library is available on GitHub (\href{github.com/gfnorg/torchgfn}{https://github.com/GFNOrg/torchgfn}) and on pypi (\href{pypi.org/project/torchgfn/}{https://pypi.org/project/torchgfn/}).
\end{abstract}

\vspace{-3.5mm}
\section{Introduction}
\label{sub:intro}
Generative Flow Networks~\citep[GFlowNets, GFNs;][]{bengio2021flow,bengio2023gflownet} are probabilistic models over discrete or continuous sample spaces with a compositional structure. They are also stochastic sequential samplers that generate objects from a target distribution, which is given by its unnormalized probability mass function $R$,  referred to as the reward function. 

The rapid growth of GFN research has led to several open-source implementations. While this signals a healthy ecosystem, the majority of the existing open source code is bespoke for a given project, and existing libraries are often tailored for specific domains (\textit{e.g.}, molecular generation\footnote{ \href{https://github.com/recursionpharma/gflownet}{github.com/recursionpharma/gflownet}}), full applications designed around complex experimental frameworks\footnote{ \href{https://github.com/alexhernandezgarcia/gflownet}{github.com/alexhernandezgarcia/gflownet}, used for \textit{e.g.}, \citet{ai4science2023crystal}.}, or are more focused on performance than extensibility or ease of use \footnote{ \href{https://github.com/d-tiapkin/gfnx}{github.com/d-tiapkin/gfnx}, a JAX-based library supporting discrete environments. }. 

We introduce \texttt{torchgfn}, the first library to offer a general-purpose, highly modular, and user-centric PyTorch-based~\citep{pytorch} toolkit designed from the ground up for extensibility and modification. Its design philosophy, centered on the strict separation of concerns, empowers researchers not only to replicate existing work but, more importantly, to accelerate the next generation of GFN research across a wide spectrum of applications. It decouples the environment definition, the sampling process, and the parametrization used for the GFN loss. The library aims to introduce new users to GFNs and their continuous variants~\citep{lahlou2023continuous,sendera2024improved,zhang2023diffusion}, and facilitate the development of new algorithms. 

The library is shipped with many example environments in the \texttt{Gym} which capture various GFN use-cases. Some examples include 1) a discrete environment where all states are terminating states (\textit{e.g.}, \texttt{HyperGrid}~\citep{bengio2021flow}); 2) discrete environments where all trajectories are of the same length, but only some states are terminating (\textit{e.g.}, \texttt{DiscreteEBM}); 3) Simple autoregressive settings (\textit{e.g.}, \texttt{BitSequence}); 4) continuous environments with state-dependent action spaces (\textit{e.g.}, \texttt{Box}); 5) discrete sampling of graphs (\textit{e.g.}, \texttt{BayesianStructure} \citep{deleu2022bayesian} and \texttt{Ring}); and 6) diffusion-based samplers for continuous distributions (\textit{e.g.}, \texttt{DiffusionSampling}~\citep{sendera2024improved}). This list cannot be exhaustive as \texttt{Gym} will be actively developed by maintainers and the community as \texttt{torchgfn} is adopted into various application domains. These examples help users learn the theory of GFNs, reproduce published environments, illustrate the proper use of the library, and provide examples of how a user may extend base classes for specific use-cases.

\vspace{-1mm}
\begin{figure}[h]
    \centering
    \includegraphics[width=0.9\linewidth]{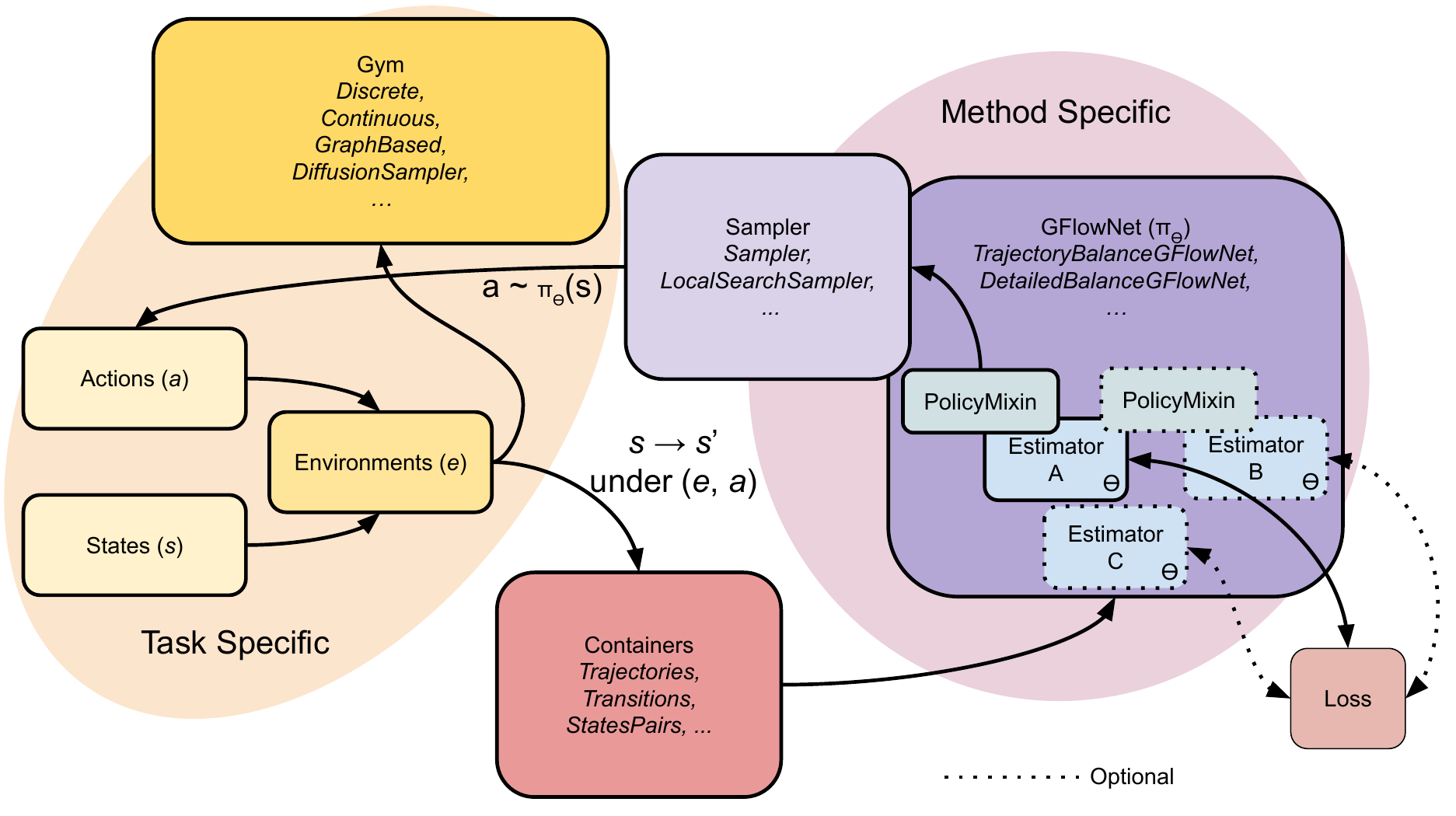}
    \caption{\small{A schematic of the modular \texttt{torchgfn} repository structure. \texttt{States} and  \texttt{Actions} are top-level abstractions used to interface between the stateless \texttt{Env}s. \texttt{Containers} are generic objects which hold \texttt{States}-\texttt{Actions} sequences drawn from \texttt{Samplers}. These interface (via a \texttt{PolicyMixin}) directly with the parametrization of a  \texttt{GFlowNet} subclass (\textit{e.g.}, \texttt{TrajectoryBalanceGFlowNet}) implementing a specific training objective. Parametrization is handled by one or more \texttt{Estimator}s (\texttt{pytorch} \texttt{nn.Module} wrappers) which represent the learned function approximators (\textit{e.g.}, policies $\pi$, state flows), trained using the \texttt{GFlowNet}'s loss. A set of \texttt{Env}s is made available through the \texttt{Gym} module.}}
    \label{fig:codeschematic}
    \vspace{-5mm}
\end{figure}

\section{Library Architecture}

The conceptual graph of the library hierarchy is shown in Figure \ref{fig:codeschematic} (see appendix Figure \ref{fig:codehierarchy} for a detailed breakdown of the full dependency structure of the library). At a high level, the Markov decision process (MDP) defined over a directed acyclic graph (DAG) is implemented by the interaction of \texttt{States} and \texttt{Actions} via a stateless \texttt{Env} (environment), which produces new \texttt{States} instances. As these elements are intimately tied to the training task, they are defined by the user in one location within the \texttt{Env} itself. \texttt{State}-\texttt{Action} \texttt{Containers} interact with \texttt{GFlowNet}s parameterized by \texttt{Estimators} to sample actions (via a \texttt{Sampler}) which explore the DAG. The \texttt{Estimators} are trained using the \texttt{GFlowNet}'s \texttt{.loss()} method, and the interface between these estimators and the sampler is mediated by a \texttt{PolicyMixin}, enabling generic samplers to work with any kind of estimator architecture. A full treatment on each library element is available in appendix \ref{apx:codebase-components}, with up-to-date documentation available in the online documentation\footnote{\href{https://torchgfn.readthedocs.io/en/latest/}{https://torchgfn.readthedocs.io/en/latest/}}. There are also many working examples in the repo tutorials\footnote{\href{https://github.com/GFNOrg/torchgfn/tree/master/tutorials}{https://github.com/GFNOrg/torchgfn/tree/master/tutorials}}.

\textbf{Environments}  $e$ are \textit{stateless} objects with a step function which produces a state transition $s'$ from $s$ given an action $a$, \textit{i.e.}, $s \rightarrow s'$ under $(e, a)$. Their stateless construction facilitates easy scaling. 
\textbf{States} $s$ are the primitive building blocks for GFlowNet training objects, which contain state-level metadata such as \textit{masks} for disallowing undesirable or impossible state transitions. They can be of any datatype, but must be castable to a format compatible with \texttt{pytorch} and/or \texttt{torch\_geometric}. \textbf{Actions} $a$ represent the internal actions of an agent building a compositional object. \textbf{Containers} wrap \texttt{States} and \texttt{Actions} during the sampling process in the correct format for a given \texttt{GFlowNet}. \textbf{Preprocessors} are responsible for casting the datatypes contained in \texttt{States} into those compatible with your defined \texttt{Estimators}. \textbf{Estimators} are wrappers for \texttt{nn.Modules} or other \texttt{pytorch} module-like abstractions which transform preprocessed \texttt{States} into distributions of actions. \textbf{Samplers} define how actions are sampled at each state, relying on an Estimator to produce a distribution to be sampled from. \textbf{GFlowNets} encapsulate the required \texttt{Preprocessor}, \texttt{Estimator}, \texttt{Sampler}, and loss-related logic to produce a trained sampler. Aside from \texttt{nn.Module}-based support, we also provide support for \textit{sampling graph objects} using \texttt{torch\_geometric}, see \cref{apx:sampling-graphs} for more details.

This shows how to build and train a Trajectory Balanced-based GFN \citep{malkin2022trajectory}:

\begin{python}
env = HyperGrid(ndim=4, height=8, reward_fn_kwargs={"R0": 0.01})
prep = KHotPreprocessor(ndim=env.ndim, height=env.height)
module_PF = MLP(input_dim=prep.output_dim, output_dim=env.n_actions) 
module_PB = MLP(input_dim=prep.output_dim, output_dim=env.n_actions - 1)
gfn = TBGFlowNet(  # Define the GFlowNet.
    pf=DiscretePolicyEstimator(
        module_PF, env.n_actions, is_backward=False, preprocessor=prep), 
    pb=DiscretePolicyEstimator(
        module_PB, env.n_actions, is_backward=True, preprocessor=prep),
)
sampler = Sampler(estimator=gfn.pf)
optimizer = torch.optim.Adam(gfn.parameters(), lr=1e-3)

for i in (pbar := tqdm(range(1000))):  # Training.
    trajectories = sampler.sample_trajectories(env=env, n=16)
    optimizer.zero_grad()
    loss = gfn.loss(env, trajectories)
    loss.backward(); optimizer.step()
\end{python}

\vspace{-1.5mm}
To instead train a SubTBGFlowNet, simply change the parameterization of the GFlowNet:

{\small
\begin{python}
module_logF = MLP(input_dim=prep.output_dim, output_dim=1)
gfn = SubTBGFlowNet(  # Define the SubTrajectory GFlowNet.
    pf=DiscretePolicyEstimator(
        module_PF, env.n_actions, is_backward=False, preprocessor=prep), 
    pb=DiscretePolicyEstimator(
        module_PB, env.n_actions, is_backward=True, preprocessor=prep),
    logF=ScalarEstimator(module=module_logF, preprocessor=prep),
)
\end{python}
}

\vspace{-5mm}
\subsection{Design Philosophy in Contrast with Other Libraries}
\label{sec:comparison}

\texttt{torchgfn} is designed as a domain-agnostic, general-purpose library, providing out-of-the-box support for discrete, continuous, and graph-based environments, under a unified API optimized for extensibility. This versatility makes it an ideal platform for foundational research on the GFlowNet framework itself -- innovations can be tested on well-understood environments where execution speed and optimization are not the top priority. \texttt{recursionpharma/gflownet}~\citep[hereafter \texttt{recursion}]{recursionpharma2023} is explicitly specialized for graph and molecular generation, leveraging \texttt{torch\_geometric} and \texttt{networkx} as its core data representations, while \texttt{alexhernandezgarcia/gflownet}~\citep[hereafter \texttt{gflownet}]{hernandezgarcia2023} is most deeply engineered for complex scientific applications such as crystal generation in addition to standard environments, and the JAX-based \texttt{gfnx}~\citep{tiapkin2025gfnx} provides discrete environments and fewer abstractions, in exchange for a JIT-compiled sampling loop with substantial speed advantages.  While powerful for their respective domains, each library's specialization creates barriers for adaptation to certain problem types and fundamental GFlowNets methods (see \cref{tab:comparison}). The core abstractions of \texttt{torchgfn} (\cref{apx:torchgfn-environments,apx:torchgfn-states,apx:torchgfn-actions,apx:torchgfn-containers,apx:torchgfn-estimators,apx:torchgfn-samplers,apx:gflownets-and-losses}) are compositional objects that can be modified by users to cleanly implement specific or novel methods. For example, the GFN training objective itself is an interchangeable \texttt{GFlowNet} object, clearly separating the algorithm from the neural network models (\texttt{Estimator}s). The \texttt{Sampler} and \texttt{Estimator} logic are also disentangled via the swappable \texttt{PolicyMixin} interface. 


\vspace{-5mm}
\begin{table}[h]
\centering
\caption{High-level comparison of GFlowNet libraries.}
\label{tab:comparison}
\footnotesize

\begin{tabularx}{\textwidth}{l|>{\columncolor{gray!10}}X|X|X|X}

\toprule
\textbf{Library} & \textbf{\texttt{torchgfn}} & \textbf{\texttt{recursion}} & \textbf{\texttt{gflownet}} & \textbf{\texttt{gfnx}}  \\
                 & Pytorch & Pytorch & Pytorch & JAX \\
\midrule
\textbf{Philosophy} & Modular \& Decoupled & Graph (Molecule)-Focused & Environment-Centric & Low-Level \& Fast \\
\midrule
\textbf{Envs} & Stateless & Stateless & Stateful & Stateless \\
\midrule
\textbf{Focus} & General-Purpose & Graph \& Molecule & General-Purpose & Discrete Envs \\
\midrule
\textbf{Batching} &
\tiny{\texttt{torch.Tensor} \& \texttt{torch\_geometric.Data}} & \tiny{\texttt{torch\_geometric.Batch}} & \tiny{Custom \texttt{Batch} Class} & \tiny{\texttt{jax.vmap}} \\
\midrule
\textbf{Loss Impl.} & Classes & Classes & Methods in Agent & User-Defined \\
\bottomrule
\end{tabularx}
\end{table}

\vspace{-2mm}
Compared with alternatives, \texttt{torchgfn} inhabits the middle ground between performance and usability, aiming to serve the needs of users who want to get ideas off the ground quickly and are less concerned with the fastest training time (Table \ref{tab:benchmark_small}, see Appendix \ref{sec:benchmark} for full benchmark results).

\begin{wraptable}{r}{0.55\textwidth}
\vspace{-15mm} 
\centering
\footnotesize
\caption{Per-iteration times averaged across 3 seeds on Hypergrid (batch size 128).}
\label{tab:benchmark_small}
\begin{tabular}{llrrrrr|rr}
\toprule
Scenario & Library & Time (ms) & Mem (MB) & Params \\
  \midrule
  Small & gflownet & 141.4 & 24 & 70,915 \\
  Small & \textbf{torchgfn} & 30.8 & 24 & 71,430 \\
  Small & gfnx & 2.5 & 182 & 71,429 \\
  \midrule
  Medium & gflownet & 223.0 & 26 & 83,717 \\
  Medium & \textbf{torchgfn} & 77.8 & 43 & 84,746 \\
  Medium & gfnx & 5.9 & 220 & 84,745 \\
  \midrule
  Large & gflownet & 364.7 & 37 & 100,101 \\
  Large & \textbf{torchgfn} & 113.3 & 55 & 101,130 \\
  Large & gfnx & 10.8 & 361 & 101,129 \\
\bottomrule
\end{tabular}
\vspace{-5mm}
\end{wraptable}

\vspace{-4mm}
\section{Future Work \& Conclusions}

Some limitations we intend to address during future development include 1) simplifying the extensible, modular design of \texttt{States}, \texttt{Actions}, and \texttt{Env}s to make environment definition easier and be more compatible with \texttt{torch.compile}, 
2) extend the library with real-world tasks containing more complex state spaces relevant to the language model and AI for science communities, 
3) supporting hierarchical sampling natively by storing sampling metadata in \texttt{Trajectories}\footnote{Currently, users would need to subclass the \texttt{PolicyMixin} and \texttt{Trajectories} classes for such use-cases.} 
4) support for multi-objective \textit{meta-environment}s, 
and 5) the inclusion of reinforcement learning baselines. 

In sum, \texttt{torchgfn} employs a modular architecture that empowers researchers to develop and test new GFN algorithms as self-contained, importable classes, which we believe addresses an unmet need in the GFN community. 
It represents a significant refinement of GFN building blocks into a set of well-documented and tested abstractions, simplifying experimentation and improving the readability and reproducibility. Our project already has many external contributors and we are in the process of fostering a dynamic developer community for GFNs with proper contribution standards\footnote{\hyperlink{github.com/GFNOrg/torchgfn/blob/master/.github/CONTRIBUTING.md}{github.com/GFNOrg/torchgfn/blob/master/.github/CONTRIBUTING.md}}. We intend for it to become the go-to community standard for foundational research. 

\newpage

\acks{The authors would like to extend a great thanks to the members of the community who contributed valuable pull requests: Alexandre Larouche, Mike Arpaia, Idriss Malek, Abhijith Sharma, Ali Parviz, Etienne Collin, Emir Ceyani, Joseph Bunao, Aya Laajil, Sanchit Misra, Chirayu Haryan, \& Inel Djafar. We would like to thank Mo Tiwari, Edward Hu, Tristan Deleu, Daniel Jenson, Eric Elmoznino, Nikolay Malkin, Emmanuel Bengio, Alex Hernandez-Garcia \& Anja Surina for invaluable discussions and feedback. This work was supported by Intel, Samsung, and the Ministry of Economy and Innovation (MEI) Quebéc.}

\vskip 0.2in
\bibliography{sample}

@article{bengio2021flow,
    title={Flow Network based Generative Models for Non-Iterative Diverse Candidate Generation},
    author={Emmanuel Bengio and Moksh Jain and Maksym Korablyov and Doina Precup and Yoshua Bengio},
    journal={Advances in Neural Information Processing Systems (NeurIPS)},
    year={2021},
}

@article{laajil2025curriculum,
  title={Curriculum-Augmented {GF}low{N}ets For m{RNA} Sequence Generation},
  author={Laajil, Aya and Shtanchaev, Abduragim and Muhammad, Sajan and Moulines, Eric and Lahlou, Salem},
  journal={arXiv preprint arXiv:2510.03811},
  year={2025}
}

@misc{tensordict2023,
      title={Torch{RL}: A data-driven decision-making library for {PyTorch}},
      author={Albert Bou and Matteo Bettini and Sebastian Dittert and Vikash Kumar and Shagun Sodhani and Xiaomeng Yang and Gianni De Fabritiis and Vincent Moens},
      year={2023},
      eprint={2306.00577},
      archivePrefix={arXiv},
      primaryClass={cs.LG}
}

@misc{recursionpharma2023,
  author = {Bengio, Emmanuel and others},
  title = {recursionpharma/gflownet},
  year = {2023},
  publisher = {GitHub},
  journal = {GitHub repository},
  howpublished = {\url{https://github.com/recursionpharma/gflownet}},
}

@misc{hernandezgarcia2023,
  author = {Hernandez-Garcia, Alex and others},
  title = {alexhernandezgarcia/gflownet},
  year = {2023},
  publisher = {GitHub},
  journal = {GitHub repository},
  howpublished = {\url{https://github.com/alexhernandezgarcia/gflownet}},
}

@article{zhang2022scheduling,
  title     = {Robust Scheduling with {GF}low{N}ets},
  author    = {David W. Zhang and Corrado Rainone and M. Peschl and R. Bondesan},
  journal   = {International Conference on Learning Representations (ICLR)},
  year      = {2023},
}

@article{bengio2023gflownet,
  title={Gflownet foundations},
  author={Bengio, Yoshua and Lahlou, Salem and Deleu, Tristan and Hu, Edward J and Tiwari, Mo and Bengio, Emmanuel},
  journal={Journal of Machine Learning Research (JMLR)},
  volume={24},
  number={210},
  pages={1--55},
  year={2023}
}

@article{malkin2022trajectory,
  title={Trajectory Balance: Improved Credit Assignment in {GF}low{N}ets},
  author={Malkin, Nikolay and Jain, Moksh and Bengio, Emmanuel and Sun, Chen and Bengio, Yoshua},
  journal={Neural Information Processing Systems (NeurIPS)},
  year={2022}
}

@article{sendera2024improved,
  title     = {Improved off-policy training of diffusion samplers},
  author    = {Marcin Sendera and Minsu Kim and Sarthak Mittal and Pablo Lemos and Luca Scimeca and Jarrid Rector-Brooks and Alexandre Adam and Y. Bengio and Nikolay Malkin},
  journal   = {Neural Information Processing Systems (NeurIPS)},
  year      = {2024},
}

@article{pytorch,
title = {Py{T}orch: An Imperative Style, High-Performance Deep Learning Library},
author = {Paszke, Adam and Gross, Sam and Massa, Francisco and Lerer, Adam and Bradbury, James and Chanan, Gregory and Killeen, Trevor and Lin, Zeming and Gimelshein, Natalia and Antiga, Luca and Desmaison, Alban and Kopf, Andreas and Yang, Edward and DeVito, Zachary and Raison, Martin and Tejani, Alykhan and Chilamkurthy, Sasank and Steiner, Benoit and Fang, Lu and Bai, Junjie and Chintala, Soumith},
journal = {Advances in Neural Information Processing Systems (NeurIPS)},
year = {2019},
}

@article{lahlou2023continuous,
	title        = {A theory of continuous generative flow networks
},
	author    = {Salem Lahlou and T. Deleu and Pablo Lemos and Dinghuai Zhang and Alexandra Volokhova and Alex Hernández-García and L'ena N'ehale Ezzine and Y. Bengio and Nikolay Malkin},
	year         = 2023,
	journal={International Conference on Machine Learning (ICML)},
}

@article{madan2022learning,
  title={Learning {GF}low{N}ets from partial episodes for improved convergence and stability},
  author={Madan, Kanika and Rector-Brooks, Jarrid and Korablyov, Maksym and Bengio, Emmanuel and Jain, Moksh and Nica, Andrei and Bosc, Tom and Bengio, Yoshua and Malkin, Nikolay},
  journal={International Conference on Machine Learning (ICML)},
  year={2023}
}

@article{deleu2022bayesian,
  title={Bayesian structure learning with generative flow networks},
  author={Deleu, Tristan and G{\'o}is, Ant{\'o}nio and Emezue, Chris and Rankawat, Mansi and Lacoste-Julien, Simon and Bauer, Stefan and Bengio, Yoshua},
  journal={Uncertainty in Artificial Intelligence (UAI)},
  year={2022},
}

@article{kim2023local,
  title={Local search gflownets},
  author={Kim, Minsu and Yun, Taeyoung and Bengio, Emmanuel and Zhang, Dinghuai and Bengio, Yoshua and Ahn, Sungsoo and Park, Jinkyoo},
  journal={International Conference on Learning Representations (ICLR)},
  year={2024}
}

@article{ai4science2023crystal,
  title={Crystal-gfn: sampling crystals with desirable properties and constraints},
  author={AI4Science, Mila and Hernandez-Garcia, Alex and Duval, Alexandre and Volokhova, Alexandra and Bengio, Yoshua and Sharma, Divya and Carrier, Pierre Luc and Benabed, Yasmine and Koziarski, Micha{\l} and Schmidt, Victor},
  journal={arXiv preprint arXiv:2310.04925},
  year={2023}
}

@article{pan2023better,
  title={Better Training of {GF}low{N}ets with Local Credit and Incomplete Trajectories},
  author={Pan, Ling and Malkin, Nikolay and Zhang, Dinghuai and Bengio, Yoshua},
  journal={International Conference on Machine Learning (ICML)},
  year={2023},
}

@article{harris2020array,
  title={Array programming with {NumPy}},
  author={Harris, Charles R and Millman, K Jarrod and Van Der Walt, St{\'e}fan J and Gommers, Ralf and Virtanen, Pauli and Cournapeau, David and Wieser, Eric and Taylor, Julian and Berg, Sebastian and Smith, Nathaniel J and others},
  journal={nature},
  volume={585},
  number={7825},
  pages={357--362},
  year={2020},
  publisher={Nature Publishing Group UK London}
}

@article{zhang2023diffusion,
  title={Diffusion generative flow samplers: Improving learning signals through partial trajectory optimization},
  author={Zhang, Dinghuai and Chen, Ricky TQ and Liu, Cheng-Hao and Courville, Aaron and Bengio, Yoshua},
  journal={International Conference on Machine Learning (ICML)},
  year={2024}
}

@book{sutton1998reinforcement,
  title={Reinforcement learning: An introduction},
  author={Sutton, Richard S and Barto, Andrew G and others},
  volume={1},
  number={1},
  year={1998},
  publisher={MIT press Cambridge}
}

@article{tiapkin2025gfnx,
  title={gfnx: Fast and Scalable Library for {G}enerative {F}low {N}etworks in {JAX}},
  author={Tiapkin, Daniil and Agarkov, Artem and Morozov, Nikita and Maksimov, Ian and Tsyganov, Askar and Gritsaev, Timofei and Samsonov, Sergey},
  journal={arXiv preprint arXiv:2511.16592},
  year={2025}
}

\newpage

\appendix

\section{Library Structure}

\begin{sidewaysfigure}[ht]
    \centering
    \includegraphics[width=1\linewidth]{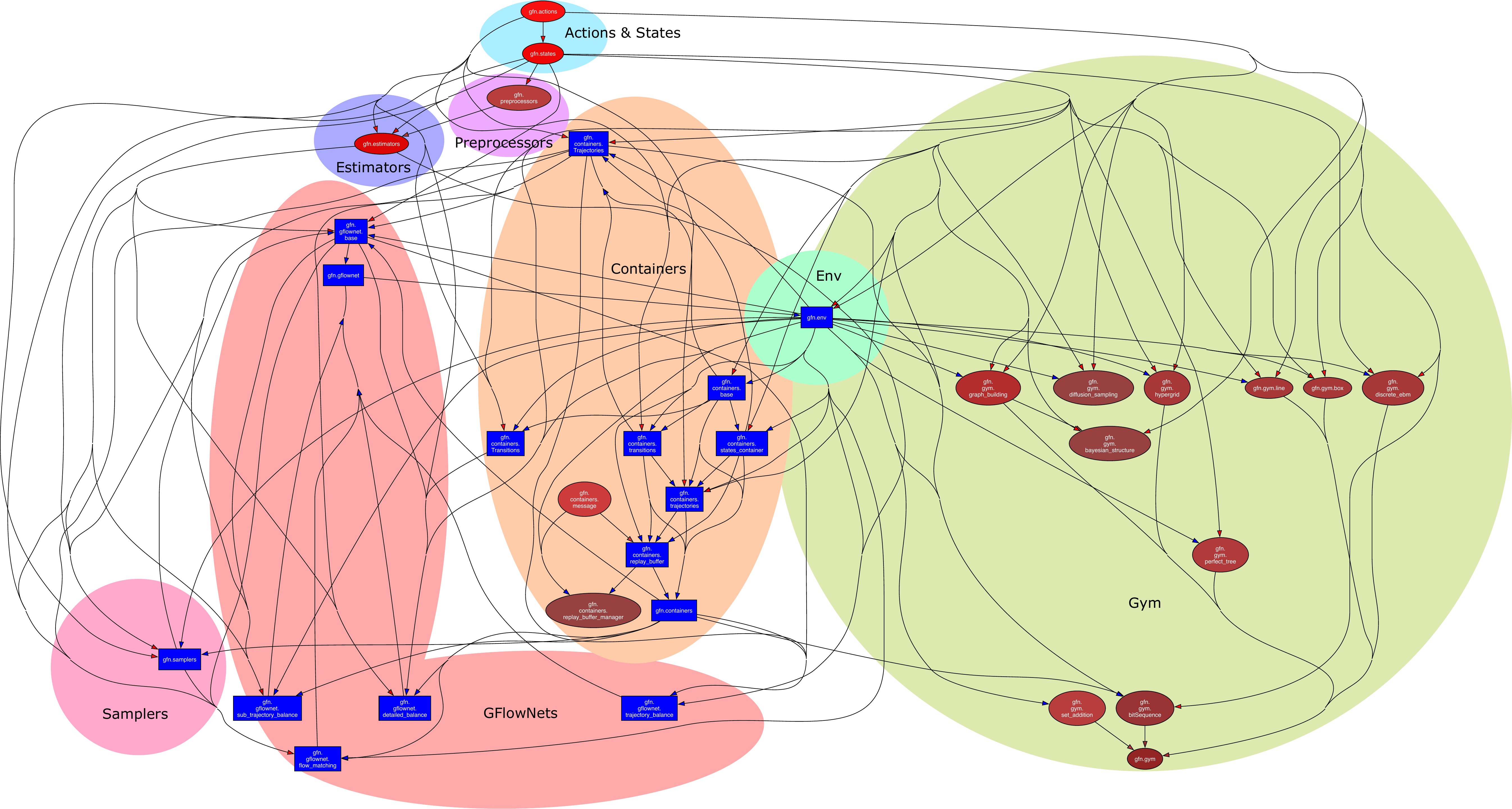}
    \caption{\texttt{torchgfn} dependency graph as of \texttt{v2.3.0}. \texttt{States} and  \texttt{Actions} are top-level abstractions used to interface between the stateless \texttt{Env} and a \texttt{GFlowNet}. \texttt{Containers} wrap \texttt{States}-\texttt{Actions} sequences as either \texttt{Transition} or \texttt{Trajectory}-level information to be used by the remainder of the library. They're drawn from \texttt{Samplers} via \texttt{Estimator}s (wrappers for \texttt{pytorch}'s \texttt{nn.Module}) representing function approximators (\textit{e.g.}, policies, state flows) housed within the \texttt{GFlowNet} subclass (\textit{e.g.}, \texttt{TrajectoryBalanceGFlowNet}). The interface between \texttt{Samplers} and \texttt{Estimators} is mediated by a \texttt{PolicyMixin}. The \texttt{GFlowNet} subclass contains all loss-relevant logic. \texttt{Environment}s are made available through the \texttt{Gym} module.}
    \label{fig:codehierarchy}
\end{sidewaysfigure}

A full breakdown of the internal library dependency structure can be seen (as of \texttt{v2.3.0}) in Figure \ref{fig:codehierarchy}. The high-level library structure is as follows:

\begin{itemize}
    \item \texttt{actions.py}: Objects representing actions. 
    \item \texttt{states.py}: Objects representing states.
    \item \texttt{env.py}: Base classes for all environments.
    \item \texttt{containers/}: Objects that hold collections of \texttt{States}, \texttt{Actions}, \texttt{Estimator} outputs, and other auxiliary metadata for training a GFlowNet. This includes \texttt{ReplayBuffer}s.
    \item \texttt{preprocessors.py}: functions for converting \texttt{States} into objects that \texttt{Estimator}s can consume. 
    \item \texttt{estimators.py}: \texttt{nn.Module} wrappers, in the form of \texttt{Estimators} and \texttt{PolicyMixin}s, which encompass the learnable parameters of a GFlowNet and their interface with \texttt{Sampler}s and log probability calculations.
    \item \texttt{samplers.py}: Classes encompassing sampling logic.
    \item \texttt{gflownet/}: Classes that encapsulate the full parameterization of a trainable GFlowNet, including the loss and sampling procedure.
    \item \texttt{gym/}: A collection of environments.
    \item \texttt{utils/}: misc utility functions.

\end{itemize}
\section{Codebase Components}
\label{apx:codebase-components}
\subsection{Environments}
\label{apx:torchgfn-environments}

In \texttt{torchgfn}, environments $e$ are stateless objects with a step function which given an action $a$ and state $s$ produces a transformed state $s'$, \textit{i.e.}, $s \rightarrow s'$ under $(e, a)$. The stateless nature of the environment is a deliberate design choice which allows for parallelism during training in large scale, multinode settings where queries to the environment (for next state computation and/or reward computation) can be broadcast among as many dedicated compute nodes as required, decoupling the computation bottlenecks potentially imposed when GFlowNet action sampling is significantly faster than environment queries, as is often the case when the environment involves complex energy functions. Environment definition requires the user to specify key components of the MDP, including a representation of the initial state $s_0$ and terminal state $s_f$ to ensure that the MDP is structured as a pointed DAG \citep{bengio2023gflownet}, a \texttt{States} class factory that contains the logic governing action masking (\textit{i.e.}, keeping track of and enforcing all allowable actions given the current state), and finally an \texttt{Actions} class factory that defines a representation of sampled actions legible to the environment. Specific environment abstractions are provided, for example, \texttt{DiscreteEnv} allows easily specifying the action space using the total number of actions as an attribute. The environment must either implement a \texttt{log\_reward()} or \texttt{reward()} method which would be called at every terminating state (\textit{i.e.}, a state with only $s_f$ as a child in the DAG) and potentially at non-terminal states if allowed by the environment and the GFlowNet is to be trained using intermediate rewards \citep{pan2023better}\footnote{Environment designers should be mindful of numerical stability issues arising from the scale of reward values.}.

When defining an environment, besides \texttt{s0}, users can optionally define a tensor representing the sink state $s_f$, which is only used for padding incomplete trajectories. If not specified, \texttt{sf} is set to a tensor of the same shape as \texttt{s0} filled with $-\infty$.

For \texttt{DiscreteEnv}s, the user can define a \texttt{get\_states\_indices()} method that assigns a unique integer number to each state, and a \texttt{n\_states} property that returns an integer representing the number of states (excluding $s_f$) in the environment. The function \texttt{get\_terminating\_states\_indices()} can also be implemented and serves the purpose of uniquely identifying terminating states of the environment, which is helpful for tabular \texttt{Estimator}s. Other properties and functions can also be implemented, such as the \texttt{log\_partition} or the \texttt{true\_dist} properties.

\subsection{States}
\label{apx:torchgfn-states}

\textbf{States} are the primitive building blocks for GFlowNet training objects, such as transitions and trajectories, on which losses operate. The provided abstract \texttt{States} class must be subclassed for each environment to define $s_0$, $s_f$, and the states shape for a single batch element. This enables the \texttt{States} instance to contain \texttt{Environment}-specific state information, allowing the environment itself to remain stateless. A \texttt{States} object is a collection of states, typically stored as a \texttt{torch.tensor}-based batch for efficient processing. In cases where the environment requires heterogeneous data, such as graphs with varying numbers of nodes and edges alongside global properties, \texttt{torchgfn} leverages \texttt{numpy.Array} \citep{harris2020array} and \texttt{tensordict}~\citep{tensordict2023}, a library from the PyTorch ecosystem. \texttt{Numpy} arrays are used for efficient management of \texttt{torch\_geometric} objects, and \texttt{TensorDict} is a dictionary-like container that holds tensors of different shapes and types while allowing for unified batching and device management operations, leading to cleaner and more efficient code which efficiently manages multiple external pytorch libraries towards the goal of training a GFlowNet.

\textbf{Masks:} Not all actions are always possible at all states -- this constraint is handled by the \texttt{State}'s mask property. \texttt{DiscreteStates} and \texttt{GraphStates} objects (\texttt{States} specialized for environments where states are represented as discrete-variable tensors and graphs, respectively) have both \texttt{forward\_masks} and \texttt{backward\_masks}, specifying the actions allowed in each state and the actions that could have produced that state, respectively. Masks can be implemented either as an attribute of the \texttt{States} class, which is kept updated via the \texttt{update\_masks} method defined in the environment, or as a property, allowing them to be generated on demand at each state. While one can technically train a GFlowNet without masks, the number of invalid trajectories sampled during training may be so great that the loss will not converge in reasonable wall-clock time, so it is almost always a practical necessity and important component of state implementation.

\textbf{DataTypes:} Notably, the data contained within a \texttt{States} object can take any form (\textit{e.g.}, strings, \texttt{numpy} arrays), but \texttt{torchgfn} only provides explicit support for some forms which allow for efficient GFN training leveraging \texttt{pytorch}, including \texttt{torch.Tensor}s and \texttt{torch\_geometric}'s graph \texttt{Data} objects. To handle alternative representations, the user must provide a custom \texttt{Preprocessor} to the \texttt{Estimator} when defining your \texttt{GFlowNet} which transforms these data structures into batched tensors.

\subsection{Actions}
\label{apx:torchgfn-actions}

\textbf{Actions} represent internal actions of an agent building a compositional object: they correspond to transitions $s \rightarrow s'$. An abstract \texttt{Actions} class is provided. While it is automatically subclassed for simple discrete environments, it must be manually subclassed in other cases because \texttt{Actions} require knowledge of both the state representation, and therefore the environment. In the \texttt{DiscreteEnv} case, an action is simply an integer representing an index between $0$ and $n_{actions} - 1$. Additionally, environments that allow for early termination of trajectories require a \texttt{exit\_action} tensor corresponding to the termination action ($[n_{actions} - 1]$ for discrete environments), unless all trajectories have a fixed length. For discrete environments, the action set $\{0, \dots, n_{actions} - 1\}$ contains also a $(n_{actions})$-th \textit{exit} or \textit{terminate} action (\textit{i.e.}, $s \rightarrow s_f$), corresponding to the index $n_{actions} - 1$. 

\FloatBarrier
\FloatBarrier

\subsection{Containers}
\label{apx:torchgfn-containers}

\textbf{Containers} wrap \texttt{States} and \texttt{Actions} during the sampling process -- they contain all elements required for training a GFlowNet in the form of abstractions such as \texttt{Transitions} or \texttt{Trajectories}. For example, these can hold the outputs of estimators during the sampling of a trajectory to facilitate the calculation of log probabilities when training a model using off-policy methods, as well as other metadata that might be useful for calculating the loss under a particular parameterization.

Containers are collections of \texttt{States}, along with other information, such as reward values or densities $p(s' \mid s)$. Three containers are available:

\begin{itemize}
    \item Transitions, representing a batch of transitions $s \rightarrow s'$.
    \item Trajectories, representing a batch of complete trajectories $s_0 \rightarrow s_1 \rightarrow \dots \rightarrow s_n \rightarrow s_f$.
    \item StatesContainer, representing a batch of  states (for example, parents and children i.e., $(s_p, s_c)$, for the flow matching loss \cite{bengio2021flow}.
\end{itemize}

These containers can either be instantiated using a \texttt{States} object or can be initialized as empty containers that can be populated on the fly, allowing the usage of the \texttt{ReplayBuffer} class, to enable experience replay while training your GFlowNet off policy to stabilize training with uncorrelated trajectories \citep{sutton1998reinforcement}. They inherit from the base \texttt{Container} class, indicating some helpful methods. In most cases, one needs to sample complete trajectories. From a batch of trajectories, a batch of states and a batch of transitions can be defined using \texttt{Trajectories.to\_states()} and \texttt{Trajectories.to\_transitions()}, in order to train GFlowNets with losses that are edge-decomposable or state-decomposable. These exclude meaningless transitions and dummy states that were added to the batch of trajectories to allow for efficient batching. 

\subsection{Estimators}
\label{apx:torchgfn-estimators}

Training GFlowNets using a \texttt{pytorch} optimizer requires at least one \texttt{Estimator} (an abstract subclass of \texttt{torch.nn.Module}). In addition to the usual \texttt{forward} method, \texttt{Estimator}s implement other useful attributes that interface with the rest of the \texttt{torchgfn} library, for example, ensuring \texttt{torch.nn.Module} outputs have the correct output dimensions for the action space, or the required logic to convert module outputs into probability distributions. 

For all \texttt{Estimator}s, the \texttt{forward} function accepts a \texttt{States} object. Neural network estimators require tensors in a particular format, and therefore one may need to define a \texttt{Preprocessor} object as part of the environment that transforms the \texttt{States} representation into something compatible with the \texttt{Estimator} in question (\cref{apx:torchgfn-environments}). 

As an example, a \texttt{DiscretePolicyEstimator} is an \texttt{Estimator} that can be used to define the policies $P_F(. \mid s)$ and $P_B(. \mid s)$ for discrete environments. At initialization, when \texttt{is\_backward=False}, the required output dimension is \texttt{n\_actions}, and when \texttt{is\_backward=True}, it is \texttt{n\_actions - 1} \footnote{There is no exit action for backward policies.}. These numbers represent the logits of a categorical distribution. The corresponding \texttt{to\_probability\_distribution()} function transforms the logits by masking illegal actions (according to the forward or backward masks contained in the \texttt{States} instance), and returns a categorical distribution. Masking is accomplished by setting the corresponding logit to $-\infty$. The function also includes exploration parameters, in order to define a tempered version of $P_F$, or a mixture of $P_F$ with a uniform distribution. Other simple examples for \textit{discrete environments} include the \texttt{Tabular} module, which implements a lookup table that can be used instead of a neural network, and a \texttt{UniformPB} module, which implements a uniform backward policy. 

For \textit{non-discrete environments}, the user needs to specify their own policies $P_F$ and $P_B$. Each module should accept a batch of \texttt{States} and return batched parameters of \texttt{torch.Distribution}s. The distribution required depends on the environment and may also depend on the previous state itself. Our tutorials and examples contain the various implementation details -- for example, in the \texttt{Box} environment, the forward policy has support either on a quarter disk or an arc-circle, such that the angle and the radius (for the quarter disk part) are scaled samples from a mixture of Beta distributions\footnote{The provided \texttt{Box} example shows an intricate scenario, and user-defined environments are not expected to need this much detail in general.}. The \texttt{to\_probability\_distribution()} function handles the conversion of the parameter outputs to an actual batched \texttt{Distribution} object that implements at least the \texttt{sample()} and \texttt{log\_prob()} functions.

\textbf{Recurrent Policies:} For sequence models that maintain hidden state (\textit{e.g.}, RNN, or autoregressive Transformers), \texttt{torchgfn} provides \texttt{RecurrentDiscretePolicyEstimator}, which extends the standard policy interface with explicit carry management. The estimator's forward pass accepts both states and a carry dictionary, returning updated logits and the next carry state. A \texttt{PolicyMixin} architecture allows these recurrent estimators to integrate seamlessly with the library's sampling and probability calculation utilities through a unified rollout API, handling per-step artifact tracking and context management automatically.

\subsection{Samplers}
\label{apx:torchgfn-samplers}
\texttt{Sampler} objects define how actions are sampled at each state. They require an \texttt{Estimator} that implements the \texttt{to\_probability\_distribution()} method. They also include a method \texttt{sample\_trajectories()}
that samples a batch of trajectories starting from a given set of initial states or $s_0$. For off-policy sampling, the parameters of \texttt{to\_probability\_distribution()} can be directly passed when initializing the \texttt{Sampler}. Samplers can follow any logic, and users can define their own, which can be passed to \texttt{GFlowNet} to enable new functionality. For example, see the included \texttt{LocalSearchSampler}, which facilitates efficient local exploration by iteratively destroying and re-sampling from high-reward trajectories \citep{kim2023local}.

To accommodate diverse policy architectures, \texttt{torchgfn} uses a \texttt{PolicyMixin} framework that provides a uniform rollout API for both vectorized (stateless) and non-vectorized (recurrent) estimators. This mixin exposes methods for initializing rollout context, computing distributions, and evaluating log-probabilities, allowing the \texttt{Sampler} and probability calculators to remain agnostic to whether an estimator maintains internal state. Per-step artifacts such as log-probabilities and estimator outputs are automatically tracked in a \texttt{RolloutContext}, which is managed by the mixin throughout trajectory generation.

\subsection{GFlowNets and Losses}
\label{apx:gflownets-and-losses}
GFlowNets can be trained with different losses, each requiring a different parametrization, so these are available as a unified \texttt{GFlowNet} object in the library. A key architectural feature of \texttt{torchgfn} is the treatment of the GFlowNet loss function as an interchangeable object. A researcher can define a set of \texttt{Estimator}s for policies and state flows, and then seamlessly switch the training objective by simply passing these modules to a different \texttt{GFlowNet} class instance (\textit{e.g.}, \texttt{TrajectoryBalanceGFlowNet}, \texttt{DetailedBalanceGFlowNet}, \textit{etc.}). This compositional approach promotes faster and cleaner experimentation. It is a meta-\texttt{Estimator} that includes one or multiple \texttt{Estimator}s, at least one of which implements a \texttt{to\_probability\_distribution()} function. They must also implement a \texttt{loss()} function that takes either states, transitions, or trajectories as input, depending on the loss. The implemented losses are the flow matching loss \citep{bengio2021flow}, the detailed balance loss \citep{bengio2023gflownet} and its modified variant \citep{deleu2022bayesian}, the trajectory balance loss \citep{malkin2022trajectory}, the sub-trajectory balance loss  \citep{madan2022learning}, and the log partition variance loss \citep{zhang2022scheduling}.

\section{Sampling Graph Objects}
\label{apx:sampling-graphs}
Sampling a graph is possible in \texttt{torchgfn} using \texttt{torch.Tensor} representations (\textit{e.g.}, building adjacency matrices), which can be suboptimal and quite limited in the general case, or by using the included \texttt{GraphEnv} and \texttt{GraphStates}, which support core operations from \texttt{torch\_geometric} in the typical \texttt{torchgfn} training setup. This allows for \texttt{Estimators} to be built using GNN-based \texttt{Estimator}, for example, \texttt{GraphEdgeActionGNN}, which assumes a fixed number of nodes and samples only edges between them. The action space for graph building can be whichever subset of valid graph actions required for a task, enabling maximum flexibility for environment designers requiring graph-based states. We have multiple examples in the library, \textit{e.g.}, \texttt{train\_graph\_ring.py} and \texttt{train\_bayesian\_structure.py}, exemplifying their use, and will continue to build upon this functionality.

\section{What's New in v2}

The v2 release represents a substantial maturation of the library in general, and introduces some major improvements over the initial release from 2022:

\begin{itemize}
    \item \textbf{Graph generation support:} First-class support for graph generation using graph-based states via \texttt{torch\_geometric} integration into our \texttt{GraphStates} class.
    \item \textbf{Conditional GFlowNets:} Support for conditional generation through conditioning tensors passed to estimators, as used in \citet{laajil2025curriculum} for instance.
    \item \textbf{Custom samplers:} Support for user-defined sampling strategies, exemplified by the \texttt{LocalSearchSampler} for efficient local exploration~\citep{kim2023local}.
    \item \textbf{Recurrent policies:} A \texttt{PolicyMixin} architecture that unifies stateless and recurrent estimators through a common rollout API, enabling sequence models (RNN, LSTM, GRU, Transformers) to maintain hidden state during trajectory generation.
    \item \textbf{Diffusion sampling:} Support for diffusion-based samplers through the \texttt{DiffusionSampling} environment, enabling GFlowNet training for continuous distributions~\citep{sendera2024improved}.
    \item \textbf{Rich ecosystem:} Expanded and tested tutorials alongside mature online documentation at \href{https://torchgfn.readthedocs.io}{https://torchgfn.readthedocs.io}.
\end{itemize}

\section{Developer Experience and Project Maturity}
\label{sec:maturity}
For an open-source library, the quality of the developer experience is critical for adoption and long-term success. \texttt{torchgfn} provides a \textit{rich learning ecosystem}, including documentation on ReadTheDocs, a quickstart guide, detailed tutorials with runnable scripts and interactive Jupyter notebooks. This comprehensive approach significantly lowers the barrier to entry for new users. The library employs a \textit{mature testing suite} using \texttt{pytest} and \textit{enforces code quality} with \texttt{pre-commit} hooks. A continuous integration (CI) pipeline managed by GitHub Actions automatically runs tests, validates tutorials, and handles publishing to PyPI. This level of automation and quality assurance is important for a production-ready library. \texttt{torchgfn} utilizes \texttt{poetry} for dependency management via a \texttt{pyproject.toml} file. This modern approach ensures reproducible environments and simplifies package distribution, aligning with current best practices in the Python ecosystem.

\subsection{Greatly Improved Modularity to Support an Expanded Algorithmic Toolkit}

Recent development work has focused on greatly improving the modularity of the library to support the introduction of new features. The \texttt{PolicyMixin} framework exemplifies this modular design: it provides a uniform interface for both vectorized (stateless) and non-vectorized (recurrent) policies through a small rollout API that handles distribution computation, log-probability calculation, and per-step artifact tracking. This allows the same \texttt{Sampler} and probability utilities to drive different estimator families -- discrete, graph-based, conditional, and recurrent -- without bespoke integration code. Similarly, the introduction of conditional \texttt{Estimators}, \texttt{LocalSearchSampler}~\citep{kim2023local}, and diffusion sampling support required only local additions or modifications to the library. We believe this modularity will be a major asset going forward as the community develops new methods or applications utilizing GFNs.

\subsection{A Rich Hub for Tutorials and Reproducibility}

The \texttt{torchgfn} library also contains a large volume of interactive tutorials which are tested using continuous integration, ensuring that they continue to function as the library evolves\footnote{\href{https://github.com/GFNOrg/torchgfn/tree/master/tutorials/notebooks}{github.com/GFNOrg/torchgfn/tree/master/tutorials/notebooks}}, and examples which serve as a rich resource for learning the basics of GFlowNets, as well as reproductions of recently published results, \textit{e.g.}, \citep{kim2023local,deleu2022bayesian,lahlou2023continuous,malkin2022trajectory,zhang2022scheduling}\footnote{\href{https://github.com/GFNOrg/torchgfn/tree/master/tutorials/examples}{github.com/GFNOrg/torchgfn/tree/master/tutorials/examples}}. These examples are also tested under continuous integration. We envision this library supporting a centralized and standardized collection of GFlowNet-related reproductions and code-based learning resources of general use to the community as the suite of tasks grows to encompass various applications in the literature.

\subsection{Performance Comparison with Alternative Libraries}
\label{sec:benchmark}

We compared \texttt{torchgfn}, \texttt{gflownet}, and \texttt{gfnx} across nine scenario configurations spanning four environment families as of February 10th, 2026. We ran HyperGrid at three scales: small (dimension=2, height=8, state space=$8^2 = 64$), medium (dimensions=4, height=16, state space=$16^4 = 65,536$), and large (dimension=4, height=32, state space= $32^4 = 1,048,576$); Ising at two scales: 6x6 and 10x10; Box2D with either a learned or uniform backward policy; and BitSequence small (sequence size=8, word size=1, modes=2) and medium (sequence size=8, word size=2, modes=4). All scenarios use Trajectory Balance loss. Across all runs, the GFN policy architecture was matched as closely as possible, with some tolerance for library-specific decisions (for example, the handling of exit actions) leading to a standard deviation of parameter count changes $\pm 10.2\%$. Each run begins with 50 warmup iterations that are excluded from timing. This allows PyTorch CUDA kernels to cache and JAX functions to JIT-compile, ensuring subsequent measurements reflect steady-state performance. Each of the 100 timed iterations is measured independently using \texttt{time.perf\_counter()}, with explicit device synchronization before and after each iteration (\texttt{torch.cuda.synchronize()} for PyTorch; \texttt{jax.block\_until\_ready()} for JAX) to eliminate asynchronous execution artifacts. Per-phase timing (sampling, loss computation, backpropagation, optimizer step) is recorded where possible; \texttt{gfnx} reports only total step time as its JIT-compiled training step cannot be instrumented without overhead. Each scenario is run at batch sizes 32, 128, and 512 to characterize scaling behavior. \texttt{gflownet} is skipped for Ising $10x10$ at batch size 512 due to prohibitive $\mathcal{O}$(batch $\times$ action space $\times$ trajectory length) scaling in its action indexing. All experiments are repeated across 3 random seeds (0, 1, 2). We report the mean iteration time across seeds, with standard deviation capturing seed-to-seed variability. For each (scenario, batch\_size, library, seed) combination we record: mean and standard deviation of per-iteration wall-clock time, per-phase timing breakdowns, peak device memory allocation, and total trainable parameter count, which are shown in Table \ref{tab:benchmark_detail}. Log partition function estimates were checked before and after training as a sanity check that optimization executed correctly.

\begin{figure}[h]
    \vspace{-1cm}
    \centering
    \includegraphics[width=0.75\linewidth]{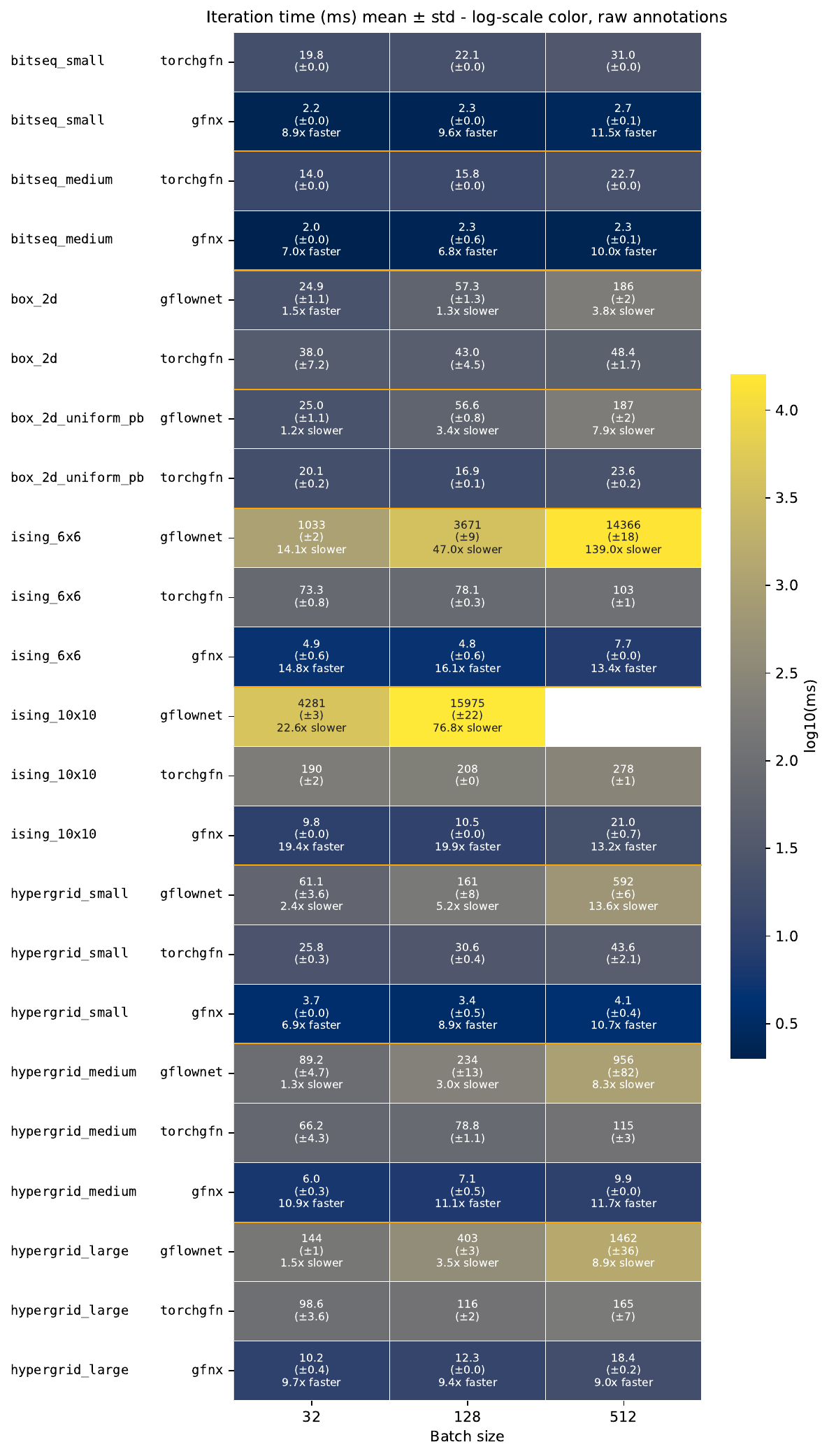}
    \caption{Comparison of the \texttt{torchgfn} library per-iteration time (ms) against the \texttt{gflownet} and \texttt{gfnx} library, across 100 iterations with a 50 iteration burn-in, with variance across 3 seeds. \texttt{Ising 10x10} is not shown for \texttt{gflownet} due to untenable computation time. Colormap is presented in log scale, speedups or slowdowns are computed relative to \texttt{torchgfn}.}
    \label{fig:benchmark}
\end{figure}

In Figure \ref{fig:benchmark} a general trend emerges: \texttt{torchgfn} occupies a comfortable middle ground between \texttt{gflownet} and \texttt{gfnx}, with runtimes roughly $2-10\times$ faster than the former and $10-20\times$ slower than the latter. With respect to \texttt{gflownet}, runtime comparisons were heavily dominated by the batch size, trajectory length, and action space of the environment, which is most noticeable in our experiments in the Ising environments. The \texttt{gflownet} library has lower fixed overhead than \texttt{torchgfn} leading to superior performance at small batch sizes, but multiple design decisions of the library prevent the user from leveraging tensor vectorization along the batch dimension and cause runtimes to scale with the batch size. There are also some operations in $\mathcal{O}(batch\_size \times action\_space)$ time which can reduce performance with large action spaces. On the other hand, \texttt{gfnx} is able to leverage JIT compilation, providing substantial speedups over \texttt{torchgfn} at all batch sizes. This process highlighted to us multiple design decisions in \texttt{torchgfn} which are incompatible with using \texttt{torch.compile}, such as the structure of our \texttt{States} and \texttt{Actions} classes. We plan to resolve these in a series of improvements on the road map to v3 of the library. We do not have reason to believe that we will be able to fully close the performance gap with JAX, which as a framework is much better suited to fully compiled training loops, but we suspect we will be able to substantially close the gap while retaining the ease of development and extensibility of Pytorch.

These benchmark results can be re-computed on demand (using up-to-date commits from all libraries) using the \texttt{benchmark/} folder now included in the \texttt{torchgfn} repo. We will use this on an ongoing basis to improve the competitive performance of our library relative to it's peers.

\begin{table}[ht]
\centering
\footnotesize
\caption{Per-phase iteration time (ms, batch size 128), peak memory, and parameter count. Phase times averaged over seeds; ``---'' indicates JIT-compiled step (not decomposable into individual timings).}
\label{tab:benchmark_detail}
\begin{tabular}{llrrrrr|rr}
\toprule
Scenario & Library & Sample & Loss & Backward & Optimizer & Total & Mem (MB) & Params \\
\midrule
  bitseq\_small & torchgfn & 16.1 & 2.9 & 2.7 & 0.4 & 22.1 & 25 & 76,578 \\
  bitseq\_small & gfnx & --- & --- & --- & --- & 1.7 & 519 & 74,520 \\
  bitseq\_medium & torchgfn & 9.7 & 2.9 & 2.7 & 0.4 & 15.8 & 21 & 75,554 \\
  bitseq\_medium & gfnx & --- & --- & --- & --- & 1.4 & 519 & 72,468 \\
\midrule
  box\_2d & gflownet & 22.4 & 30.4 & 7.3 & 0.5 & 60.7 & 22 & 74,784 \\
  box\_2d & torchgfn & 23.2 & 3.4 & 14.9 & 0.8 & 42.3 & 23 & 82,495 \\
\midrule
  box\_2d\_uniform\_pb & gflownet & 22.5 & 30.5 & 7.3 & 0.5 & 60.8 & 25 & 74,784 \\
  box\_2d\_uniform\_pb & torchgfn & 9.6 & 2.2 & 6.2 & 0.4 & 18.4 & 23 & 74,528 \\
\midrule
  ising\_6x6 & gflownet & 2616.3 & 929.5 & 88.2 & 0.9 & 3635.0 & 149 & 131,367 \\
  ising\_6x6 & torchgfn & 72.9 & 2.9 & 2.0 & 0.5 & 78.3 & 56 & 112,530 \\
  ising\_6x6 & gfnx & --- & --- & --- & --- & 3.9 & 513 & 103,020 \\
\midrule
  ising\_10x10 & gflownet & 12413.1 & 3189.6 & 244.8 & 1.5 & 15849.0 & 703 & 246,119 \\
  ising\_10x10 & torchgfn & 198.3 & 5.2 & 4.0 & 0.6 & 208.0 & 166 & 194,706 \\
  ising\_10x10 & gfnx & --- & --- & --- & --- & 9.8 & 513 & 168,748 \\
\midrule
  hypergrid\_small & gflownet & 63.4 & 62.8 & 14.3 & 0.8 & 141.4 & 24 & 70,915 \\
  hypergrid\_small & torchgfn & 25.7 & 3.0 & 1.6 & 0.5 & 30.8 & 24 & 71,430 \\
  hypergrid\_small & gfnx & --- & --- & --- & --- & 2.5 & 182 & 71,429 \\
\midrule
  hypergrid\_medium & gflownet & 110.1 & 79.5 & 32.7 & 0.8 & 223.0 & 26 & 83,717 \\
  hypergrid\_medium & torchgfn & 71.6 & 3.0 & 2.7 & 0.5 & 77.8 & 43 & 84,746 \\
  hypergrid\_medium & gfnx & --- & --- & --- & --- & 5.9 & 220 & 84,745 \\
\midrule
  hypergrid\_large & gflownet & 197.0 & 122.2 & 44.8 & 0.7 & 364.7 & 37 & 100,101 \\
  hypergrid\_large & torchgfn & 105.0 & 3.0 & 4.8 & 0.5 & 113.3 & 55 & 101,130 \\
  hypergrid\_large & gfnx & --- & --- & --- & --- & 10.8 & 361 & 101,129 \\
\bottomrule
\end{tabular}
\end{table}

\end{document}